\documentclass[10pt,conference]{IEEEtran}
\IEEEoverridecommandlockouts
\usepackage{cite}
\usepackage{amsmath,amssymb,amsfonts}
\usepackage{algorithmic}
\usepackage{multirow}
\usepackage{caption}
\usepackage{booktabs} 
\usepackage{graphicx}
\usepackage{textcomp}
\usepackage{xcolor}
\usepackage{subfig}
\usepackage{algorithm}
\usepackage{colortbl}
\usepackage{xcolor}
\def\BibTeX{{\rm B\kern-.05em{\sc i\kern-.025em b}\kern-.08em
    T\kern-.1667em\lower.7ex\hbox{E}\kern-.125emX}}
\DeclareCaptionLabelFormat{upper}{\MakeUppercase{#1}~#2}
\begin{document}

\title{Asynchronous Federated Unlearning with Invariance Calibration for Medical Imaging\\

}

\author{\IEEEauthorblockN{Zhaoyuan Cai}
\IEEEauthorblockA{\textit{School of Computer Science and Engineering} \\
\textit{South China University of Technology}\\
Guangzhou, China \\
cszycai@mail.scut.edu.cn}
\and
\IEEEauthorblockN{Xinglin Zhang}
\IEEEauthorblockA{\textit{School of Computer Science and Engineering} \\
\textit{South China University of Technology}\\
Guangzhou, China \\
csxlzhang@scut.edu.cn}
}

\maketitle
\begin{abstract}
Federated Unlearning (FU) is an emerging paradigm in Federated Learning (FL) that enables participating clients to fully remove their contributions from a trained global model, driven by data protection regulations that mandate the right to be forgotten. However, existing FU methods mostly rely on synchronous coordination. This requirement forces the entire federation to halt and wait for stragglers to complete erasure, creating significant delays due to device heterogeneity. Furthermore, these methods often face the problem that the influence of erased data is merely suppressed temporarily and resurfaces during subsequent training, rather than being genuinely removed. To overcome these limitations, this paper proposes Asynchronous Federated Unlearning with Invariance Calibration (AFU-IC), a novel framework for medical imaging that decouples the erasure process from the global training workflow. This enables the target client to perform unlearning asynchronously without interrupting global training. Meanwhile, a server-side invariance calibration mechanism prevents the model from relearning the erased data. Extensive experiments on three medical benchmarks demonstrate that AFU-IC achieves unlearning efficacy and model fidelity comparable to gold-standard retraining while significantly reducing wall-clock latency compared to synchronous baselines. AFU-IC ensures efficient, compliant and reliable FL in cross-silo medical environments.
\end{abstract}


\section{Introduction}
\label{sec:intro}

Federated Learning (FL) has become a key enabling technology for collaborative medical image analysis, allowing distinct medical institutions to jointly train a robust global model without sharing raw patient data~\cite{mcmahan2017communication, Li2025Efficient,li2025achieving}. However, once a global model has been trained, it becomes difficult to remove the contribution of a specific client, which conflicts with data protection regulations—such as GDPR~\cite{voigt2017eu} and CCPA~\cite{pardau2018california}—that mandate the right to be forgotten. In clinical practice, a hospital may revoke data usage rights due to expired consent or data correction, necessitating the complete removal of its contribution from the global model. To address this issue, Federated Unlearning (FU) has recently emerged as a new paradigm that enables the selective erasure of specific clients’ data contributions from trained federated models~\cite{nguyen2025survey}.

Existing FU methods generally fall into partition-based, model-manipulation, and optimization-based categories~\cite{liu2021federaser,wu2022federated,wang2022federated,halimi2022federated,gu2024unlearning}. However, applying these approaches to medical federations faces two critical challenges:

\begin{enumerate}
    \item \textbf{Synchronous Bottleneck.} 
    Most unlearning protocols operate within a synchronous FL framework, requiring the server to aggregate updates from all clients simultaneously. In medical environments with varying hardware capabilities, this mechanism forces all retained clients to pause their normal operations and wait for the target client to complete its erasure. If the target client is a straggler, this effectively blocks the entire system, resulting in unacceptable service downtime. While recent works like KNOT~\cite{su2023asynchronous} introduce asynchrony to mitigate this global waiting, they rely on a clustered retraining mechanism. By restricting the rollback-and-retrain process to local clusters, KNOT still imposes high computational costs on the affected clients, failing to achieve low-latency erasure. 

    \item \textbf{Non-genuine Erasure.} 
    Optimization-based methods, such as Projected Gradient Ascent (PGA)~\cite{halimi2022federated}, primarily focus on maximizing the loss on target data. 
    However, this objective often induces a masking effect: the model tends to superficially shift its decision boundary to misclassify target samples, rather than eliminating the underlying feature representations. 
    Consequently, the model parameters remain structurally trapped within the original optimization basin. When subsequently updated by retained clients, these latent, suppressed features are easily reactivated, leading to the model reverting~\cite{pan2025federated}.
\end{enumerate}

These challenges give rise to a critical question: How to design an FU mechanism that overcomes the efficiency bottlenecks of synchronization-based schemes, while ensuring that deleted information cannot be recovered during subsequent training, thus enabling genuine unlearning?

To address this problem, we propose a novel \textbf{Asynchronous Federated Unlearning Framework with Invariance Calibration (AFU-IC)}. We adapt the reference-model constraint from previous work~\cite{halimi2022federated} to a fully asynchronous setting, allowing the target client to execute erasure independently without blocking the federation, thereby eliminating the waiting time for retained clients. Crucially, to overcome the model reverting issue of PGA-based methods, we introduce a server-side invariance calibration mechanism. Instead of merely breaking the model's performance on target data, we enforce feature-level unlearning by minimizing the KL-divergence between clean and triggered predictions. This aligns the model's behavior with that of a robustly retrained oracle, ensuring that the erasure is deep and persistent.

Our main contributions are summarized as follows:

\begin{itemize}
    \item To the best of our knowledge, we are the first to introduce an asynchronous paradigm to medical FU. By resolving the latency constraints of synchronous aggregation, our protocol allows for non-blocking, on-demand erasure execution suitable for heterogeneous medical environments.
    
    \item We address the structural inadequacy of standard gradient ascent, where erased information is merely suppressed rather than removed. 
    To resolve this, we integrate a server-side invariance calibration mechanism. This KL-based objective acts as a structural regularizer, forcing the model to mathematically decouple its decision logic from the specific feature correlations of the target client. This ensures that the erasure is permanent and robust against the recovering effects of future updates.
    
    \item Extensive experiments on medical datasets demonstrate that our approach achieves superior unlearning efficacy comparable to gold-standard retraining while maintaining high fidelity on the retained data distributions~\cite{oasis_kaggle,yang2023medmnist}. Crucially, it significantly outperforms synchronous baselines in wall-clock efficiency and ensures a stable, deep erasure that effectively resists model reverting during subsequent training phases.
\end{itemize}
\section{Related Work}
\label{sec:related_work}

\subsection{Federated Learning in Healthcare}
The classical FL algorithm FedAvg~\cite{mcmahan2017communication} relies on synchronized aggregation across participating clients, with aggregation weights determined exclusively by local dataset sizes. Although effective under ideal independent and identically distributed (IID) conditions, its performance deteriorates markedly in temporally evolving and non-IID data distributions. While advancements like FedProx~\cite{li2020federated} address optimization in non-IID settings, they focus on the learning phase. How to effectively unlearn a specific client's contribution in such highly heterogeneous and cross-silo medical environments remains an open problem, as the distinct data distribution of the target client leaves a strong imprint on the global model.

\subsection{Federated Unlearning}
FU aims to remove specific clients or data samples from a trained global model. Recent comprehensive surveys~\cite{romandini2024federated,liu2024survey} provide a structured overview of existing FU strategies. 
FedEraser~\cite{liu2021federaser} adopts a storage-intensive strategy by caching historical parameter updates and calibrating them to approximate retraining. While theoretically precise, the storage overhead is often prohibitive for large medical models. 
Model manipulation approaches leverage techniques such as Knowledge Distillation (KD)~\cite{wu2022federated,li2025frequency, li2025srkd,li2025mmt} to transfer retained knowledge to a sanitized student model, or employ parameter pruning~\cite{wang2022federated} to nullify weights associated with specific classes. Optimization-based methods, specifically PGA proposed by Halimi et al.~\cite{halimi2022federated}, directly maximize the loss on target data. Similarly, FedOSD~\cite{pan2025federated} employs orthogonal steepest descent to mitigate gradient conflicts. However, these methods suffer from a fundamental limitation: they predominantly rely on synchronous aggregation.  In heterogeneous medical cross-silo settings, this dependency dictates that the unlearning process is bounded by the slowest client, creating a severe ``straggler bottleneck.'' Our work addresses this critical efficiency gap by introducing a fully asynchronous protocol that decouples the unlearning execution from the global synchronization barrier.

\subsection{Asynchronous Federated Learning}
Asynchronous Federated Learning (AFL) is developed to address the straggler problem in cross-device and cross-silo training. FedAsync~\cite{xie2019asynchronous} allows the server to update the global model immediately upon receiving a local update, using a staleness-aware weighting function. Other works like~\cite{chen2020asynchronous} have further optimized AFL for non-IID data and resource-constrained devices.
Crucially, these works focus on the training phase (learning). In contrast, our work addresses the unlearning phase (erasing), which is fundamentally different. Unlearning is inherently more unstable than training, as it involves disrupting the model's converged knowledge structure. For instance, methods relying on gradient ascent push parameters away from the local optimum, which can easily trigger catastrophic forgetting~\cite{kirkpatrick2017overcoming}. Applying AFL techniques directly to unlearning without regularization often leads to model collapse. Our work bridges this gap by tailoring asynchronous mechanisms specifically for the unlearning objective.

\subsection{Knowledge Erasure and Preservation}
To prevent catastrophic forgetting of the main task during unlearning while ensuring the target influence is truly removed, regularization techniques are essential. EWC~\cite{kirkpatrick2017overcoming} and its variants constrain parameter updates based on Fisher Information to retain old tasks. In the context of unlearning, methods like FedRecover~\cite{zhang2023fedrecovery} and FedQUIT~\cite{fedquit2024} utilize KD primarily to align the unlearned model with the original model's behavior on retained data.

However, existing works largely neglect the structural stability of the erasure itself. Our work repurposes KD as a dual-purpose mechanism: invariance calibration. By enforcing consistency between original samples and their transformed views (containing specific non-robust features), our regularization forces the model to ignore the specific correlations targeted for unlearning. This mechanism not only preserves utility on the retained distribution but also enforces structural unlearning, preventing the model from simply shifting to an invalid parameter space and ensuring convergence towards the theoretically retrained oracle.
\section{Methodology}
\label{sec:method}
\begin{figure}[t]
    \centering
    \includegraphics[width=1.0\linewidth]{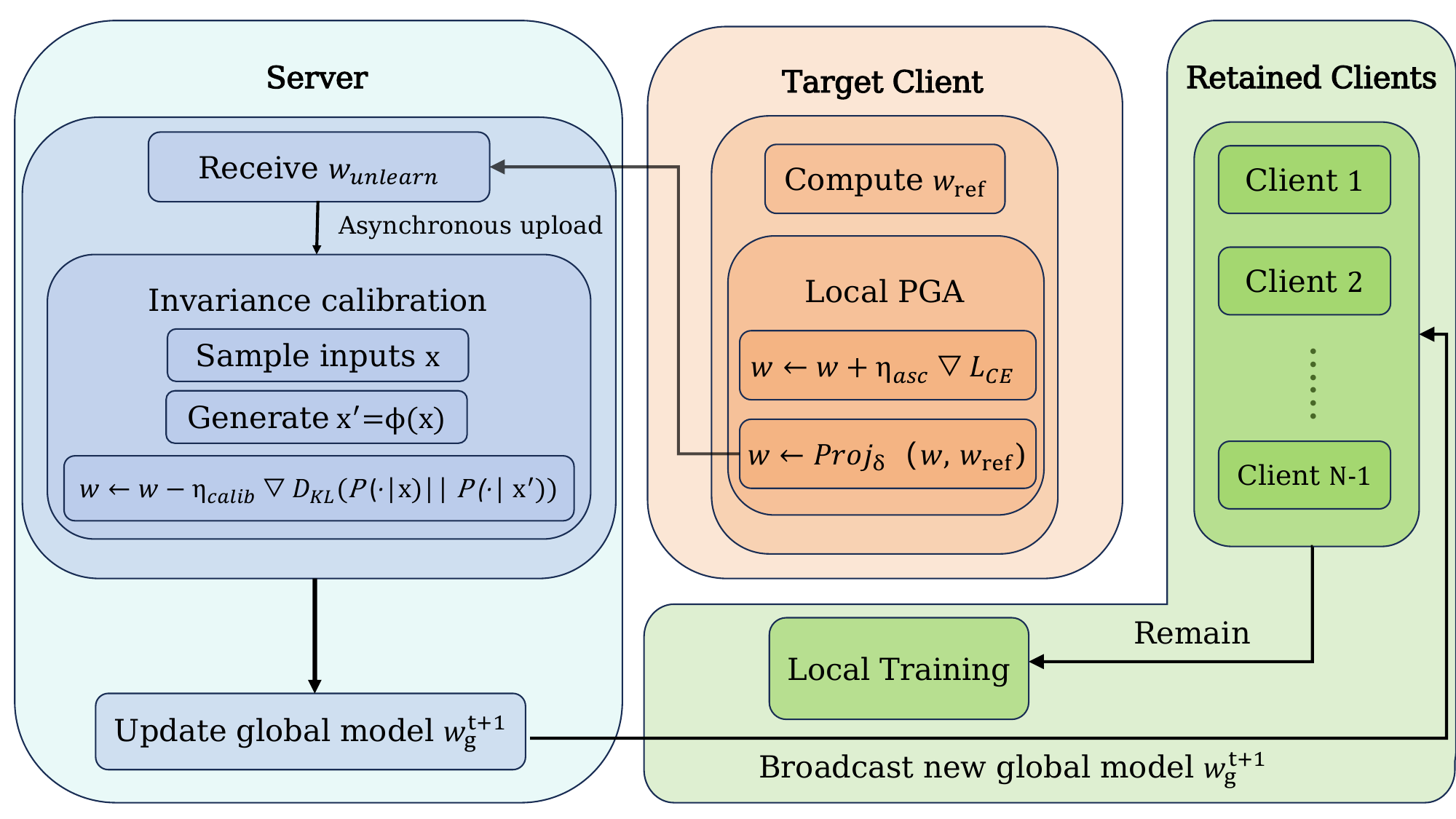}
    \caption{Overview of the AFU-IC framework.}
    \label{fig:afu_ic_framework}
\end{figure}

In this section, we introduce our proposed AFU-IC. We formulate the unlearning problem as a constrained optimization task aimed at approximating the retraining oracle. As illustrated in Figure \ref{fig:afu_ic_framework}, our method adopts a dual-phase strategy, separating the local erasure performed by the target client from a server-side structural calibration, ensuring both process efficiency and erasure robustness.

\subsection{Problem Formulation}

Consider an FL system with $N$ clients, collectively optimizing a global objective defined as:
\begin{equation}
    w^* = \arg\min_{w} \sum_{k=1}^N \mathcal{L}(w; \mathcal{D}_k).
\end{equation}
Let $w_{g}$ be the converged global model parameters containing information from the entire distributed dataset $\mathcal{D} = \bigcup_{k=1}^N \mathcal{D}_k$.
A target client $u$ requests to unlearn its specific local subset $\mathcal{D}_u$. The theoretical goal is to obtain an unlearned model $w_{unlearn}$ that perfectly approximates the retrained model, $w_{retrain}$, which is defined as the model trained from scratch on the retained datasets, i.e., $w_{retrain} = \arg\min_{w} \sum_{k \neq u} \mathcal{L}(w; \mathcal{D}_k)$.
Mathematically, the converged global model $w_{g}$ has deviated from the optimal solution of $w_{retrain}$ due to the specific influence of $\mathcal{D}_u$. Our objective is to rectify this parameter shift and realign the model's optimization trajectory with that of the retained distribution $\mathcal{D}_{retain} = \mathcal{D} \setminus \mathcal{D}_u$.

\subsection{Asynchronous Reference Approximation}

To prevent catastrophic forgetting of retained knowledge during the unlearning process, we first construct a reference model $w_{ref}$ to anchor the optimization.
Following the strategy in~\cite{halimi2022federated}, $w_{ref}$ is designed to serve as a proxy for the aggregate knowledge of the retained federation. Ideally, $w_{ref}$ is the average of models from all clients excluding the target. This can be derived efficiently by the target client locally, without server coordination, by subtracting its own accumulated contribution from the current global model state:
\begin{equation}
    w_{ref} = \frac{N}{N-N_u} w_{g}^{t} - \frac{N_u}{N-N_u} w_{u}^{t-1},
\end{equation}
where $N_{u}$ is the target client count, and $w_{u}^{t-1}$ represents the cached local model update of client $u$ from the previous round $t-1$. This $w_{ref}$ serves as the center of an $L_2$-norm ball constraint during optimization, preventing the model from drifting into invalid parameter spaces.

\subsection{Phase I: Local Asynchronous Gradient Ascent}

This phase aims to erase the influence of the target client's data through a computationally efficient, asynchronous protocol.
From an optimization perspective, the model $w_{g}$ resides in a local minimum with respect to $\mathcal{L}(w; \mathcal{D}_u)$. To approximate $w_{retrain}$, which theoretically exhibits higher loss on $\mathcal{D}_u$, we must push the parameters out of this specific optimization basin.

We employ PGA locally on the target client. Upon receiving an unlearning request, the target client $C_u$ performs the following optimization on the current global model snapshot $w_{g}^{t}$, independently of the ongoing training of other clients:
\begin{equation}
    w_{u}^{(\tau+1)} = \text{Proj}_{\delta} \left( w_{u}^{(\tau)} + \eta_{asc} \nabla \mathcal{L}(w_{u}^{(\tau)}; \mathcal{D}_u) \right),
\end{equation}
where $w_{u}^{(0)} = w_{ref}$, and the projection $\text{Proj}_{\delta}(\cdot)$ constrains the parameters within the ball $\mathcal{B}(w_{ref}, \delta)$.

{Asynchronous Protocol:} Crucially, this process is decoupled from the global synchronization barrier. Unlike synchronous approaches that freeze the federation, AFU-IC allows retained clients to continue standard training concurrently. Once the local unlearning converges to $w_{unlearn}$, it is transmitted to the server. Instead of waiting for a weighted aggregation, the server immediately adopts these parameters as the global initialization for the subsequent communication round, formally $w_{g}^{t+1} \leftarrow w_{unlearn}$. This ensures the ``forgetting" is enforced globally while the computational burden remains strictly local.

\subsection{Phase II: Server-Side Invariance Calibration}

Merely maximizing loss on $\mathcal{D}_u$ during the local phase often leads to a masking effect, where the model superficially shifts its decision boundary without removing the underlying feature representations, leading to model reverting. To ensure the model structurally decouples from the unlearned features and achieves permanent erasure, we introduce a server-side invariance calibration phase.

\subsubsection{Theoretical Motivation}
To foster deep structural erasure, we draw inspiration from the principles of robust learning and invariant risk minimization~\cite{arjovsky2019invariant}. In heterogeneous federated settings, deep models often tend to rely on client-specific {non-robust features}, rather than learning universally robust causal features.
The ideal retraining oracle, $w_{retrain}$, having never observed $\mathcal{D}_u$, does not rely on these specific shortcuts. Thus, theoretically approximating the oracle can be framed as aligning the model’s behavior to ignore these non-robust features.

\subsubsection{Calibration Objective}
To achieve this structural invariance, we need to enforce prediction consistency between the original inputs $x$ and their perturbation-augmented counterparts $x' = \Phi(x)$. 
Since the model outputs are probability distributions, the KL divergence serves as the natural metric to quantify the informational discrepancy between the model's behavior on $x$ and $x'$. 
Therefore, to mathematically enforce the consistency constraint discussed above, we formulate the calibration objective as minimizing the expected KL divergence over the available samples:
\begin{equation}
    \min_{w} \mathcal{L}_{calib}(w) = \mathbb{E}_{x} \left[ D_{KL}\left( P(\cdot|x; w) \parallel P(\cdot|x'; w) \right) \right].
    \label{eq:calib_objective}
\end{equation}
This objective functions as a structural regularization term. By forcing the distribution $P(\cdot|x'; w)$ to align with $P(\cdot|x; w)$, we effectively penalize the model for being sensitive to the non-causal variations introduced by $\Phi(\cdot)$.

\noindent\textbf{Mechanism of Structural Erasure.} 
Minimizing the objective in Eq.~\ref{eq:calib_objective} directly leads to two critical effects that ensure the robustness of the unlearning:
\begin{itemize}
    \item \textbf{Fidelity Preservation:} The objective anchors the model's predictions on augmented data to match its predictions on clean data. This regularization prevents the parameters from drifting too far into random space, thereby maintaining smooth decision boundaries on retained knowledge similar to a robustly trained oracle.
    \item \textbf{Feature Decoupling:} Since the transformation $\Phi(\cdot)$ simulates non-causal variations, minimizing the divergence forces the model to ignore these specific features. This mathematically decouples the decision logic from the unlearned, non-robust correlations, ensuring the erasure is permanent rather than superficial.
\end{itemize}
The complete procedure is summarized in Algorithm \ref{alg:method}. 
First, the target client initiates the unlearning process and computes a reference model $w_{ref}$ to initialize its local parameters (Lines 3-4). Crucially, this execution is parallel to the ongoing operations of retained clients, introducing zero blocking time.
During the asynchronous phase, the model is updated via gradient ascent (Line 6) and immediately projected into a $w_{ref}$-centered constraint ball (Line 7). 
Upon receiving the uploaded $w_{unlearn}$ (Line 9), the server immediately triggers structural calibration (Line 10). 
Over $T_{calib}$ epochs, the server utilizes augmented views $x'$ (Lines 13-14) to minimize the KL divergence between predictions (Line 15), effectively decoupling non-robust features. 
Finally, the calibrated model is broadcast as the new global state $w_{g}^{t+1}$ (Line 17).

\begin{algorithm}[t]
    \caption{AFU-IC}
    \label{alg:method}
    \begin{algorithmic}[1]
        \REQUIRE Global model $w_{g}^t$; Target client $u$ with cache $w_u^{t-1}$; Target Dataset $\mathcal{D}_u$; Hyperparams $\eta_{asc}, \eta_{calib}, \delta$.
        \ENSURE Unlearned global model $w_{g}^{t+1}$.
        \STATE \textbf{System State:} Retained clients continue standard training tasks without interruption.
        \STATE Target Client $u$ executes unlearning independently:
        \STATE \quad Compute reference $w_{ref} \leftarrow \frac{N w_{g}^{t} - N_u w_{u}^{t-1}}{N - N_u}$ 
        \STATE \quad Init $w \leftarrow w_{ref}$.
        \STATE \quad \textbf{for} epoch $\tau = 1, \dots, T_{asc}$ \textbf{do}
        \STATE \quad \quad $w \leftarrow w + \eta_{asc} \nabla \mathcal{L}_{CE}(w; \mathcal{D}_u)$
        \STATE \quad \quad $w \leftarrow \text{Proj}_{\delta}(w, w_{ref})$
        \STATE \quad \textbf{end for}
        \STATE \quad Asynchronously upload $w_{unlearn} \leftarrow w$ to Server.
        
        \STATE Server triggers calibration upon receiving $w_{unlearn}$:
        \STATE \quad Initialize $w \leftarrow w_{unlearn}$.
        \STATE \quad \textbf{for} epoch $\tau = 1, \dots, T_{calib}$ \textbf{do}
        \STATE \quad \quad Sample inputs $x$
        \STATE \quad \quad Generate augmented view $x' \leftarrow \Phi(x)$
        \STATE \quad \quad $w \leftarrow w - \eta_{calib} \nabla D_{KL}(P(\cdot|x; w) \| P(\cdot|x'; w))$
        \STATE \quad \textbf{end for}
        \STATE \quad Update global model $w_{g}^{t+1} \leftarrow w$ and broadcast.
    \end{algorithmic}
\end{algorithm}
\section{Experiment}
\label{sec:experimental}

\begin{table*}[t]
\centering
\caption{Comparison of Unlearning Efficiency, Efficacy, and Fidelity on three medical datasets. }
\label{tab:main_comparison_horizontal}

\setlength{\tabcolsep}{3pt}

\begin{tabular*}{\textwidth}{@{\extracolsep{\fill}} l ccccc ccccc ccccc @{}}
\toprule

\multirow{3}{*}{\textbf{Method}} & \multicolumn{5}{c}{\textbf{OASIS }} & \multicolumn{5}{c}{\textbf{PathMNIST}} & \multicolumn{5}{c}{\textbf{OrganAMNIST }} \\
\cmidrule(lr){2-6} \cmidrule(lr){7-11} \cmidrule(lr){12-16}

& \textbf{Eff.} & \multicolumn{2}{c}{\textbf{Efficacy}} & \multicolumn{2}{c}{\textbf{Fidelity}}
& \textbf{Eff.} & \multicolumn{2}{c}{\textbf{Efficacy}} & \multicolumn{2}{c}{\textbf{Fidelity}}
& \textbf{Eff.} & \multicolumn{2}{c}{\textbf{Efficacy}} & \multicolumn{2}{c}{\textbf{Fidelity}} \\
\cmidrule(lr){2-2} \cmidrule(lr){3-4} \cmidrule(lr){5-6}
\cmidrule(lr){7-7} \cmidrule(lr){8-9} \cmidrule(lr){10-11}
\cmidrule(lr){12-12} \cmidrule(lr){13-14} \cmidrule(lr){15-16}

& \textbf{Time(s)} & \textbf{BA(\%)} & \textbf{Gap} & \textbf{CA(\%)} & \textbf{Gap}
& \textbf{Time(s)} & \textbf{BA(\%)} & \textbf{Gap} & \textbf{CA(\%)} & \textbf{Gap}
& \textbf{Time(s)} & \textbf{BA(\%)} & \textbf{Gap} & \textbf{CA(\%)} & \textbf{Gap} \\
\midrule

Retrain
& 5,420 & 30.32 & - & 75.10 & - 
& 6,800 & 26.85 & - & 89.03 & - 
& 7,200 & 24.39 & - & 83.88 & - \\ 

PGA
& 2,850 & 90.74 & +60.42 & 78.23 & +3.13
& 1,920 & 92.10 & +65.25 & 88.42 & -0.61
& 1,100 & 98.50 & +74.11 & 90.23 & +6.35 \\

FedRecovery
& 1,280 & 88.52 & +58.20 & 83.52 & +8.42
& 1,550 & 96.57 & +69.72 & 84.56 & -4.47
& 1,720 & 72.88 & +48.49 & 82.14 & -1.74 \\

FedEraser
& 2,150 & 25.59 & -4.73 & 82.36 & +7.26
& 2,500 & 34.28 & +7.43 & 88.43 & -0.60
& 2,900 & 23.83 & -0.56 & 88.20 & +4.32 \\

FedOSD
& 1,180 & 23.80 & -6.52 & 82.17 & +7.07
& 1,450 & 32.19 & +5.34 & 91.97 & +2.94
& 1,600 & 21.97 & -2.42 & 90.82 & +6.94 \\

\textbf{AFU-IC}
& \textbf{295} & \textbf{28.25} & \textbf{-2.07} & \textbf{83.91} & \textbf{+8.81}
& \textbf{350} & \textbf{21.50} & \textbf{-5.35} & \textbf{90.81} & \textbf{+1.78}
& \textbf{380} & \textbf{22.20} & \textbf{-2.19} & \textbf{93.62} & \textbf{+9.74} \\

\bottomrule
\end{tabular*}
\end{table*}

\subsection{Experimental Setup}
\label{sec:experimental-setup}
\noindent\textbf{Implementation Details.}
We simulate a cross-silo federated learning setting with one central server and $N$ clients. By default, we set the number of clients $N=5$, and vary $N \in \{5, 10, 15, 20\}$ in the ablation study to evaluate scalability. For optimization, we employ a combination of SGD and Adam optimizers with a batch size of $128$.

\noindent\textbf{Datasets and FL Partitioning.} 
To verify the generalizability of our framework across diverse medical modalities, we evaluate AFU-IC on three heterogeneous benchmarks: 
1) {OASIS}~\cite{oasis_kaggle}: A dataset for dementia classification. 
2) {PathMNIST}~\cite{yang2023medmnist}: A dataset containing histology slides for colon pathology classification.
3) {OrganAMNIST}~\cite{yang2023medmnist}: A dataset with 2D abdominal CT scans.
To simulate realistic cross-silo heterogeneity, we partition these datasets among clients using a Dirichlet distribution $\text{Dir}(\alpha)$. We investigate two non-IID levels~\cite{hsu2019measuring}: $\alpha=1.0$ (moderate skew) and $\alpha=0.1$ (strong skew, long-tailed). Unless specified otherwise, experiments use $\alpha=1.0$.

\noindent\textbf{Backdoor Configuration.}
To rigorously evaluate the unlearning completeness, we adopt the backdoor injection strategy as a proxy \cite{halimi2022federated}. The target client poisons a subset of its local dataset by overlaying a trigger and modifying the ground-truth labels to a specific target class.

\noindent\textbf{Model and Baselines.}
We adopt a 4-layer CNN architecture. The model is composed of a feature extractor encompassing four stacked convolutional blocks and a projection head with two fully connected layers. This architecture effectively captures spatial dependencies in $128\times128$ medical images while maintaining computational efficiency. We compare our proposed {AFU-IC} against the following five baselines: 
(1) {Retrain}, which retrains the model from scratch on the remaining datasets to provide the oracle performance; 
(2) {PGA}~\cite{halimi2022federated}, which performs gradient ascent on the unlearning client's data with weight projection; 
(3) {FedRecovery}~\cite{zhang2023fedrecovery}, which utilizes historical gradient residuals to approximate the retrained model; 
(4) {FedEraser}~\cite{liu2021federaser}, which reconstructs the global model by calibrating historical local updates; 
(5) {FedOSD}~\cite{pan2025federated}, which employs orthogonal steepest descent to mitigate gradient conflicts during erasure.

\noindent\textbf{Evaluation Metrics.}
We assess the proposed framework from three critical dimensions:
\begin{itemize}
    \setlength\itemsep{0em}
    \item {Efficacy:} We measure erasure success using both behavioral and structural indicators: 
    1) {Backdoor Accuracy (BA)}: The accuracy on the poisoned test set. A sharp decline in BA implies the removal of the backdoor behavior.
    2) {Parameter Distance ($L_2$ Dist)}: The Euclidean distance between the unlearned model weights $w_{unlearn}$ and the retrain oracle $w_{retrain}$. Unlike BA, this structural metric detects the masking effect and ensures stability against reverting.
    \item {Fidelity:} Measured by the {Clean Accuracy (CA)} on the test set of the retained clients, quantifying how well the model preserves main-task performance.
    \item {Efficiency:} Evaluated by the total {wall-clock time (s)} required to complete the unlearning and post-learning process, accounting for both local computation and communication overhead.
\end{itemize}

Unless stated otherwise, experiments use $N=5$ clients and $\alpha=1.0$. Detailed ablation studies are deferred to Sec.~\ref{sec:rq4}.

\begin{figure*}[t] 
  \centering
  
  \subfloat[OASIS]{
    \label{fig:res_oasis}
    \includegraphics[width=0.3\linewidth]{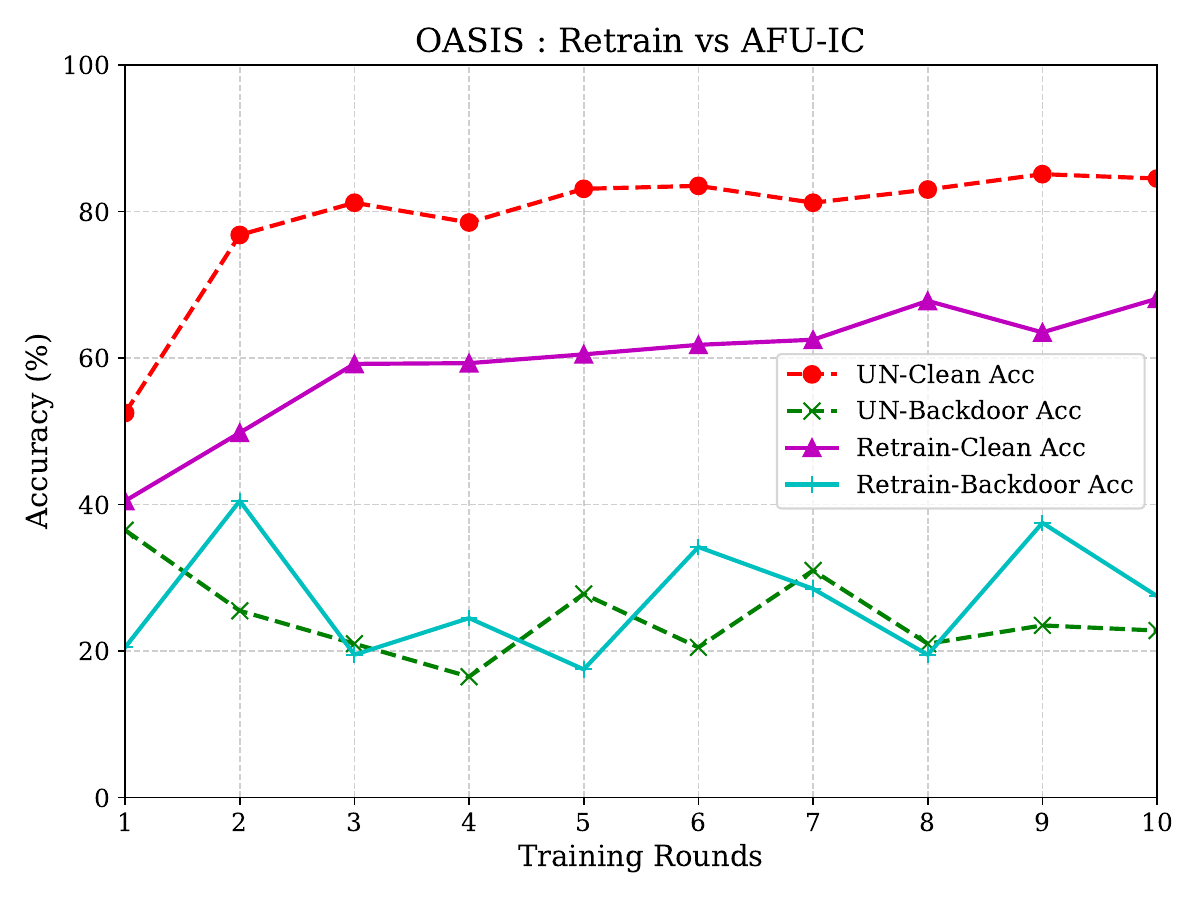}
  }
  \hfill 
  \subfloat[PathMNIST]{
    \label{fig:res_path}
    \includegraphics[width=0.3\linewidth]{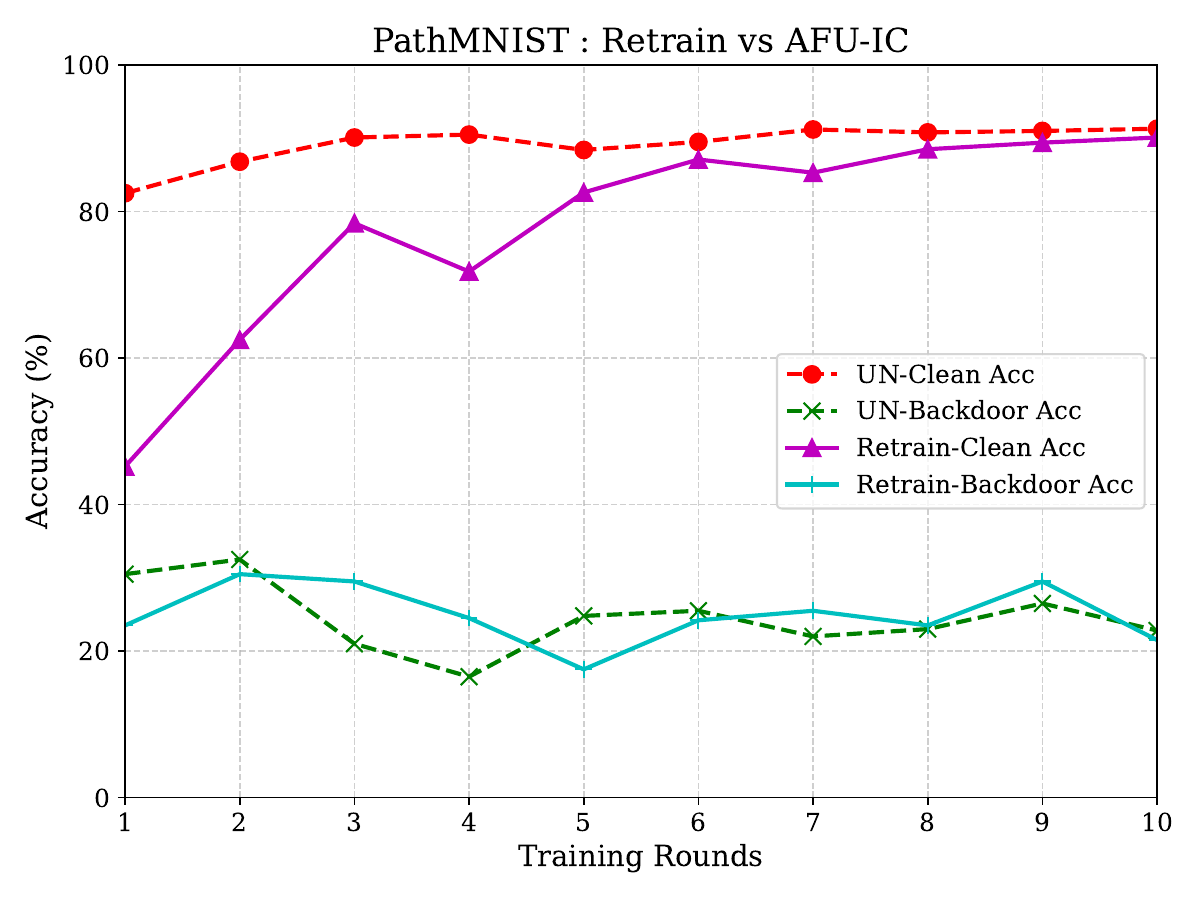}
  }
  \hfill 
  \subfloat[OrganAMNIST]{
    \label{fig:res_organ}
    \includegraphics[width=0.3\linewidth]{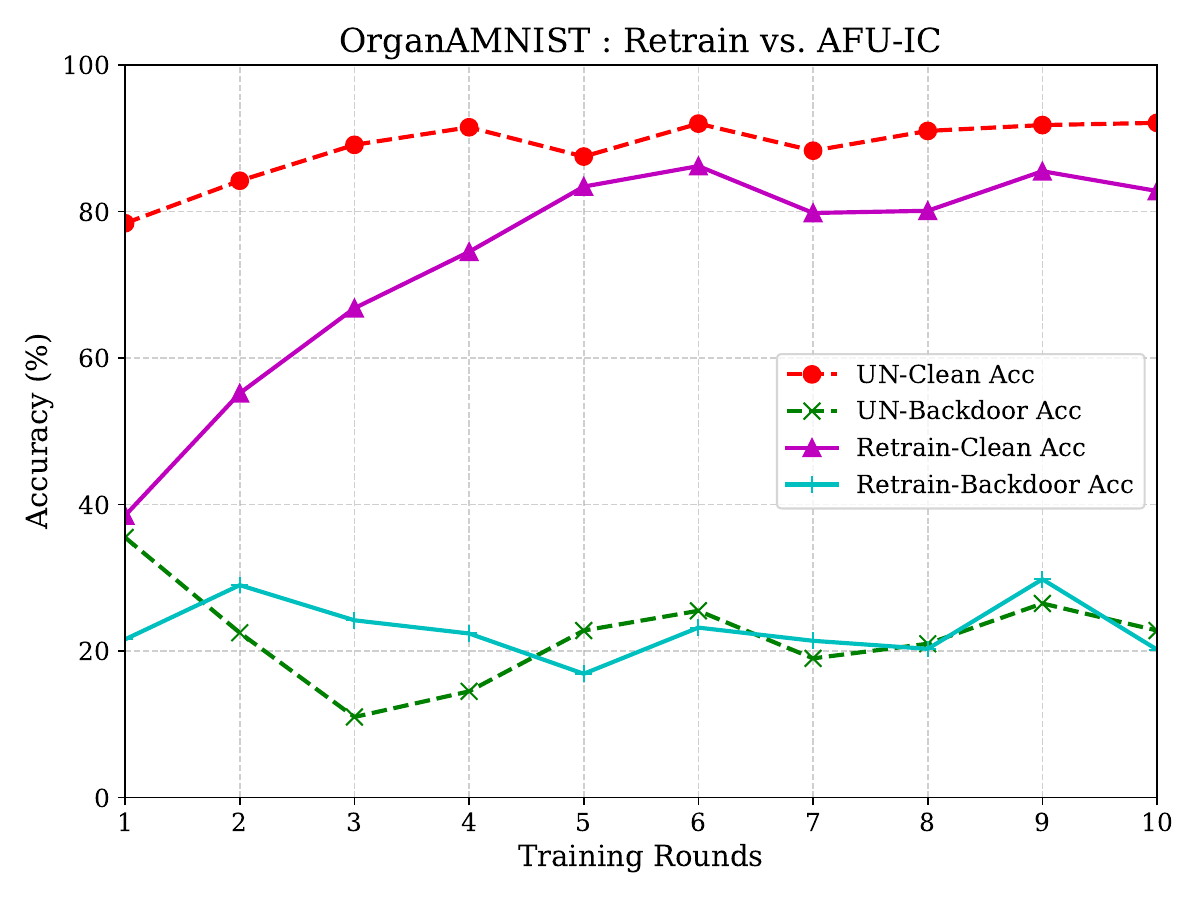}
  }
  
  \caption{Clean accuracy and backdoor accuracy of the AFU-IC and fully retrained model with respect to the number of FL post-learning rounds in each dataset for $N=5$ clients.}
  \label{fig:main_results}
\end{figure*}

\begin{figure}[t]
    \centering
    \includegraphics[width=0.95\linewidth]{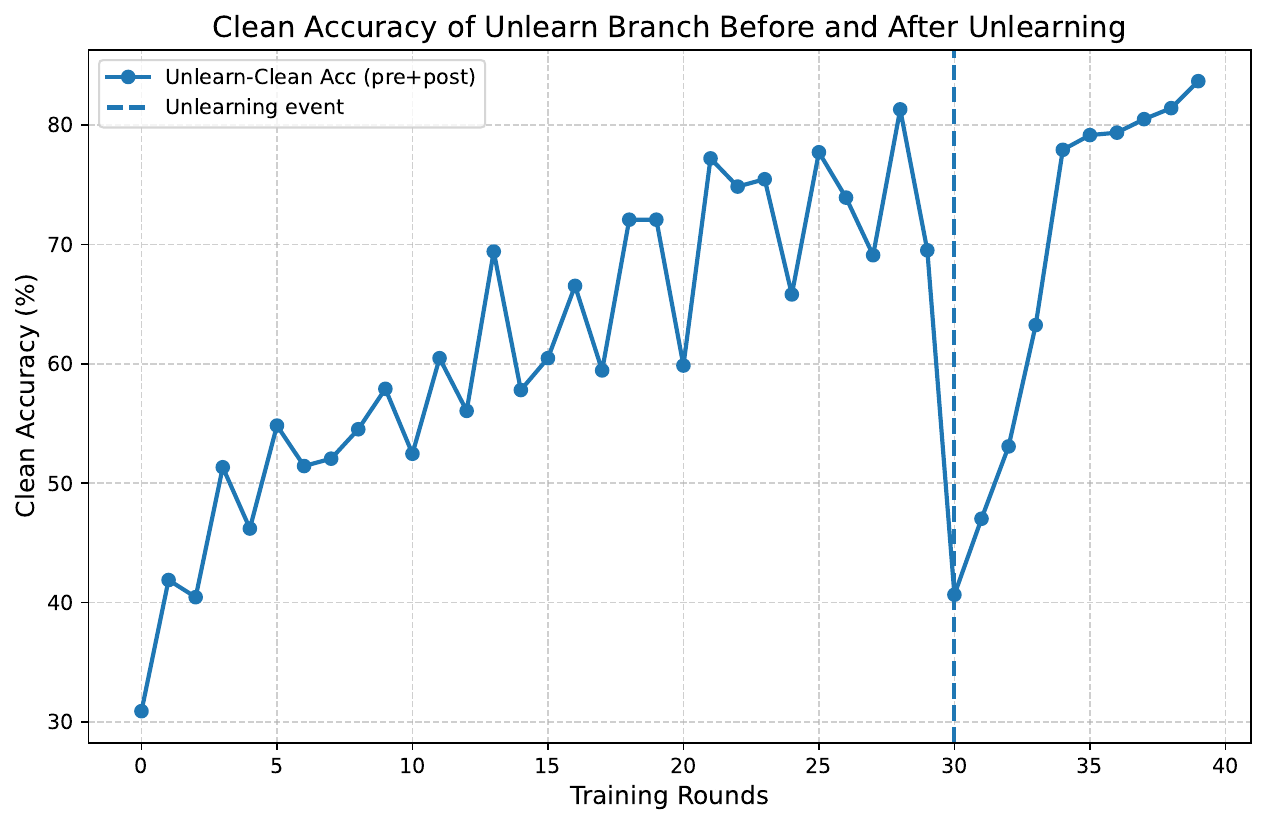}
    \caption{
        Evolution of CA throughout the federated training process.
    }
    \label{fig:unlearn_clean_curves}
\end{figure}

\begin{table}[t]
\centering
\caption{Analysis of unlearning stability on the OASIS dataset. We monitor behavioral (BA, CA) and structural ($L_2$ distance to Retrain Model) changes immediately after erasure versus after 10 post-learning rounds.}
\label{tab:reverting_l2_analysis}

\setlength{\tabcolsep}{4pt} 
\resizebox{\columnwidth}{!}{%
\begin{tabular}{l|ccc|ccc}
\toprule
\multirow{2}{*}{\textbf{Method}} & \multicolumn{3}{c|}{\textbf{Post-Unlearn (Instant)}} & \multicolumn{3}{c}{\textbf{Post-Recovery (10 Rnds)}} \\ \cmidrule(lr){2-4} \cmidrule(lr){5-7}
 & \textbf{BA}$\downarrow$ & \textbf{CA}$\uparrow$ & \textbf{$L_2$ Dist}$\downarrow$ & \textbf{BA}$\downarrow$ & \textbf{CA}$\uparrow$ & \textbf{$L_2$ Dist}$\downarrow$ \\ \midrule
 
Retrain (Oracle) & 30.32 & 65.16 & 0.00 & 30.32 & 75.17 & 0.00 \\ 

Projected-GA & 35.53 & 82.22 & 25.57 & 90.74 & 82.23 & 72.30 \\
FedRecovery & 32.13 & 81.25 & 34.49 & 88.52 & 83.52 & 64.82 \\ 

FedEraser & 25.47 & 82.19 & 30.60 & 25.59 & 82.36 & 30.62 \\
FedOSD & 23.52 & 82.01 & 20.45 & 23.80 & 82.17 & 10.48 \\ 

\textbf{AFU-IC} & \textbf{28.24} & \textbf{83.85} & \textbf{30.35} & \textbf{28.25} & \textbf{83.91} & \textbf{7.36} \\ \bottomrule
\end{tabular}%
}
\end{table}

\subsection{Unlearning Completeness (Efficacy)}
\label{sec:rq1}

\begin{table*}[t]
\centering
\caption{Comprehensive ablation study on the OASIS dataset. The \textbf{Default settings} (marked in gray) align with the updated AFU-IC results.}
\label{tab:ablation_simplified}

\setlength{\tabcolsep}{12pt} 
\renewcommand{\arraystretch}{1.1} 

\begin{tabular}{l|l|c|c|c}
\toprule
\multirow{2}{*}{\textbf{Ablation Aspect}} & \multicolumn{1}{c|}{\multirow{2}{*}{\textbf{Variant / Setting}}} & \textbf{Efficiency} & \textbf{Efficacy} & \textbf{Fidelity (Utility)} \\ \cline{3-5}
 & & \textbf{Time (s)} & \textbf{BA (\%)} $\downarrow$ & \textbf{CA (\%)} $\uparrow$ \\ \hline \hline

\multirow{3}{*}{\textbf{1. Regularization}} & $\gamma_{calib} = 0$  & 290 & 90.47 & 78.18 \\
\multirow{3}{*} & \cellcolor{gray!15}$\gamma_{calib} = 1.0$ (Ours) & \cellcolor{gray!15}300 & \cellcolor{gray!15}28.24 & \cellcolor{gray!15}83.92 \\
 & $\gamma_{calib} = 5.0$  & 320 & 22.13 & 76.47 \\ \hline

\multirow{2}{*}{\textbf{2. Mechanism}} & Synchronous FL & 1,250 & 28.18 & 84.14 \\
 & \cellcolor{gray!15}Asynchronous & \cellcolor{gray!15}300 & \cellcolor{gray!15}28.24 & \cellcolor{gray!15}83.92 \\ \hline

\multirow{2}{*}{\textbf{3. Heterogeneity}} & $\alpha = 0.1$ & 330 & 35.46 & 80.53 \\
 & \cellcolor{gray!15}$\alpha = 1.0$  & \cellcolor{gray!15}300 & \cellcolor{gray!15}28.24 & \cellcolor{gray!15}83.92 \\ \hline

\multirow{4}{*}{\textbf{4. Client Scale}} & \cellcolor{gray!15}$N = 5$  & \cellcolor{gray!15}300 & \cellcolor{gray!15}28.24 & \cellcolor{gray!15}83.92 \\
 & $N = 10$ & 350 & 26.88 & 84.25 \\
 & $N = 15$ & 410 & 25.45 & 84.41 \\
 & $N = 20$  & 460 & 24.57 & 84.54 \\ \bottomrule
\end{tabular}
\end{table*}

In this part, we evaluate the effectiveness of AFU-IC in eliminating the influence of the target client's data. We rely on the backdoor trigger as a traceable fingerprint of the target client's contribution. Successful unlearning should render the model behaviorally and structurally indistinguishable from a model that never observed the target data.

We first evaluate the unlearning efficacy by comparing the BA. As shown in Figure~\ref{fig:main_results}, the retrain baseline (Cyan curve) defines the lower bound of BA, representing the model's response to triggers when the target data is completely absent.
Crucially, our unlearned model's BA (Green curve) closely tracks the retrain baseline throughout the post-learning rounds, maintaining a small gap. 
This overlap indicates that our method has successfully scrubbed the specific patterns associated with the target client, reducing the trigger response to the same noise level as the clean baseline.

Beyond behavioral outputs, we assess the internal model updates by measuring the $L_2$ parameter distance between the unlearned model $w_{{unlearn}}$ and the retrained oracle $w_{{retrain}}$. 
Experimental results presented in Table~\ref{tab:reverting_l2_analysis} show that our method achieves a significantly smaller $L_2$ distance to the retrained model compared to baselines. 
This confirms that our AFU-IC method does not merely mask the backdoor behavior like shifting to a different local minimum but actively drives the global model parameters towards the optimization trajectory of the {retrain} oracle, thereby validating the true efficacy of unlearning.

\subsection{Utility Preservation (Fidelity)}
\label{sec:rq2}
To assess the main-task utility, we focus on the CA on retained clients.

Figure~\ref{fig:unlearn_clean_curves} demonstrates the temporal dynamics of utility. While the gradient ascent induces an expected transient drop in accuracy, the global model exhibits rapid recovery within limited communication rounds. The final CA matches the pre-unlearning performance, verifying that our KL-based regularization effectively isolates the ``forgetting" to targeted parameters without compromising the general feature extractor.

Crucially, as compared in Figure~\ref{fig:main_results}, our unlearned model consistently maintains higher CA than the retrain baseline throughout the post-learning rounds. This result highlights a significant efficiency advantage: our method preserves the valid medical knowledge acquired during Phase I, thereby avoiding the computational overhead and performance instability associated with relearning representations from random initialization.

\subsection{Efficiency Analysis}
\label{sec:rq3}

As summarized in Table~\ref{tab:main_comparison_horizontal}, AFU-IC significantly outperforms all synchronous baselines in total wall-clock time.
On the OASIS dataset, AFU-IC completes the unlearning process in just 295 seconds, representing an 18$\times$ speedup over retraining and a 4$\times$ speedup over the fastest synchronous baseline, FedOSD.
This substantial time reduction stems from our asynchronous design:
\begin{enumerate}
    \item \textbf{Zero Blocking for Retained Clients:} Unlike synchronous protocols that freeze the federation, AFU-IC eliminates the waiting time for retained clients, as they are exempted from the blocking synchronization barrier.
    \item \textbf{Decoupled Calibration:} The server-side calibration executes in parallel with ongoing local training, removing the communication bottlenecks in synchronous coordination.
\end{enumerate}
This consistent efficiency advantage across all benchmarks positions AFU-IC as a practical solution for real-time unlearning requests.

\subsection{Ablation Study}
\label{sec:rq4}

We conduct an ablation study to verify the contribution of each component in AFU-IC. The quantitative results are summarized in Table~\ref{tab:ablation_simplified}.

\noindent\textbf{Necessity of Invariance Calibration.} 
We first examine whether simple gradient ascent is sufficient for unlearning. As shown in the Regularization block, removing the calibration term ($\gamma_{calib}=0$) fails to scrub the backdoor, yielding a BA comparable to the pre-unlearning state. This indicates that without structural constraints, the optimization falls into a masking effect—shifting parameters superficially without eliminating the underlying trigger mechanism.
Incorporating the invariance constraint ($\gamma_{calib}=1.0$) effectively breaks this dependency, significantly reducing BA while preserving the main task utility. 
However, an excessively large penalty ($\gamma_{calib}=5.0$) disturbs the feature extractor, leading to a noticeable drop in CA. Thus, a moderate weight provides the optimal trade-off between erasure completeness and utility preservation.

\noindent\textbf{Efficiency of Asynchronous Mechanism.} 
To isolate the efficiency gains, we compare our framework against a synchronous variant where the server waits for all clients to complete. 
In the presence of stragglers, the synchronous baseline is severely bottlenecked, as the system pace is dictated by the slowest client. 
In contrast, our asynchronous design decouples the unlearning client from the federation, achieving an approximate $\mathbf{4\times}$ speedup in wall-clock time. 
Crucially, this acceleration does not compromise performance—both erasure efficacy and model utility remain statistically consistent with the synchronous baseline, confirming that asynchrony is an optimization for this task.

\noindent\textbf{Robustness to Heterogeneity and Scale.} 
Finally, we evaluate the method's resilience under varying data distributions and federation scales.
\begin{itemize}
    \item \textbf{Data Heterogeneity:} Under extreme non-IID settings ($\alpha=0.1$), the unlearning task becomes more challenging due to class imbalance. While we observe a marginal increase in BA compared to the moderate setting, the method successfully maintains high utility, suggesting robust feature alignment even when local distributions diverge significantly from the global prior.
    \item \textbf{Federation Scale:} We investigate scalability by varying the client size $N$. 
    With more clients, the target's relative importance decreases, which naturally helps with unlearning but increases the communication time. 
    Our results show that AFU-IC remains effective even when $N=5$, where the target contributes a large portion of the data. This proves that our method is robust regardless of whether the target client is a major or minor contributor to the federation.
\end{itemize}
\section{Conclusion}
\label{sec:conclusion}

In this work, we propose AFU-IC, a novel framework designed to overcome the limitations of existing synchronous unlearning methods. We specifically addressed the bottleneck where the entire federation is forced to halt and wait for stragglers, causing significant delays. By decoupling the erasure process from the global training workflow, our approach enables the target client to execute unlearning asynchronously without interrupting global training. Furthermore, we resolved the critical issue where the influence of erased data is merely suppressed temporarily and resurfaces during subsequent training. Through a server-side invariance calibration mechanism, we effectively prevent the model from relearning the erased contributions. Extensive experiments on three medical benchmarks confirm that AFU-IC achieves unlearning efficacy and main-task fidelity comparable to gold-standard retraining while significantly reducing wall-clock latency, providing a genuine and efficient solution for federated unlearning.
\bibliographystyle{IEEEtran} 
\bibliography{main}

@String(AAAI = {AAAI})

@inproceedings{mcmahan2017communication,
  title={Communication-efficient learning of deep networks from decentralized data},
  author={McMahan, Brendan and Moore, Eider and Ramage, Daniel and Hampson, Seth and y Arcas, Blaise Aguera},
  booktitle={Artificial intelligence and statistics},
  pages={1273--1282},
  year={2017},
  organization={PMLR}
}

@book{voigt2017eu,
  title={The EU general data protection regulation (GDPR)},
  author={Voigt, Paul and Von dem Bussche, Axel},
  year={2017},
  publisher={Springer}
}

@inproceedings{liu2021federaser,
  title={Federaser: Enabling efficient client-level data removal from federated learning models},
  author={Liu, Gaoyang and Ma, Xiaoqiang and Yang, Yang and Wang, Chen and Liu, Jiangchuan},
  booktitle={2021 IEEE/ACM 29th International Symposium on Quality of Service (IWQOS)},
  pages={1--10},
  year={2021},
  organization={IEEE}
}

@article{nguyen2025survey,
  title={A survey of machine unlearning},
  author={Nguyen, Thanh Tam and Huynh, Thanh Trung and Ren, Zhao and Nguyen, Phi Le and Liew, Alan Wee-Chung and Yin, Hongzhi and Nguyen, Quoc Viet Hung},
  journal={ACM Transactions on Intelligent Systems and Technology},
  volume={16},
  number={5},
  pages={1--46},
  year={2025},
  publisher={ACM New York, NY}
}

@article{halimi2022federated,
  title={Federated unlearning: How to efficiently erase a client in fl?},
  author={Halimi, Anisa and Kadhe, Swanand and Rawat, Ambrish and Baracaldo, Nathalie},
  journal={arXiv preprint arXiv:2207.05521},
  year={2022}
}

@article{arjovsky2019invariant,
  title={Invariant risk minimization},
  author={Arjovsky, Martin and Bottou, L{\'e}on and Gulrajani, Ishaan and Lopez-Paz, David},
  journal={arXiv preprint arXiv:1907.02893},
  year={2019}
}

@article{zhang2023fedrecovery,
  title={Fedrecovery: Differentially private machine unlearning for federated learning frameworks},
  author={Zhang, Lefeng and Zhu, Tianqing and Zhang, Haibin and Xiong, Ping and Zhou, Wanlei},
  journal={IEEE Transactions on Information Forensics and Security},
  volume={18},
  pages={4732--4746},
  year={2023},
  publisher={IEEE}
}

@article{li2020federated,
  title={Federated learning: Challenges, methods, and future directions},
  author={Li, Tian and Sahu, Anit Kumar and Talwalkar, Ameet and Smith, Virginia},
  journal={IEEE signal processing magazine},
  volume={37},
  number={3},
  pages={50--60},
  year={2020},
  publisher={IEEE}
}

@inproceedings{xie2019asynchronous,
  title={Asynchronous federated optimization},
  author={Xie, Cong and Koyejo, Sanmi and Gupta, Indranil},
  booktitle={arXiv preprint arXiv:1903.03934},
  year={2019}
}

@article{kirkpatrick2017overcoming,
  title={Overcoming catastrophic forgetting in neural networks},
  author={Kirkpatrick, James and Pascanu, Razvan and Rabinowitz, Neil and Veness, Joel and Desjardins, Guillaume and Rusu, Andrei A and Milan, Kieran and Quan, John and Ramalho, Tiago and Grabska-Barwinska, Agnieszka and others},
  journal={Proceedings of the national academy of sciences},
  volume={114},
  number={13},
  pages={3521--3526},
  year={2017},
  publisher={National Academy of Sciences}
}

@article{wu2022federated,
  title={Federated unlearning with knowledge distillation},
  author={Wu, Chen and Zhu, Sencun and Mitra, Prasenjit},
  journal={arXiv preprint arXiv:2201.09441},
  year={2022}
}

@inproceedings{pan2025federated,
  title={Federated unlearning with gradient descent and conflict mitigation},
  author={Pan, Zibin and Wang, Zhichao and Li, Chi and Zheng, Kaiyan and Wang, Boqi and Tang, Xiaoying and Zhao, Junhua},
  booktitle={Proceedings of the AAAI Conference on Artificial Intelligence},
  volume={39},
  number={19},
  pages={19804--19812},
  year={2025}
}

@article{fedquit2024,
  title={Fedquit: On-device federated unlearning via a quasi-competent virtual teacher},
  author={Mora, Alessio and Valerio, Lorenzo and Bellavista, Paolo and Passarella, Andrea},
  journal={arXiv preprint arXiv:2408.07587},
  year={2024}
}

@inproceedings{chen2020asynchronous,
  title={Asynchronous online federated learning for edge devices with non-iid data},
  author={Chen, Yujing and Ning, Yue and Slawski, Martin and Rangwala, Huzefa},
  booktitle={2020 IEEE International Conference on Big Data (Big Data)},
  pages={15--24},
  year={2020},
  organization={IEEE}
}

@article{pardau2018california,
  title={The california consumer privacy act: Towards a european-style privacy regime in the united states},
  author={Pardau, Stuart L},
  journal={J. Tech. L. \& Pol'y},
  volume={23},
  pages={68},
  year={2018},
  publisher={HeinOnline}
}

@article{liu2024survey,
  title={A survey on federated unlearning: Challenges, methods, and future directions},
  author={Liu, Ziyao and Jiang, Yu and Shen, Jiyuan and Peng, Minyi and Lam, Kwok-Yan and Yuan, Xingliang and Liu, Xiaoning},
  journal={ACM Computing Surveys},
  volume={57},
  number={1},
  pages={1--38},
  year={2024},
  publisher={ACM New York, NY}
}

@inproceedings{wang2022federated,
  title={Federated unlearning via class-discriminative pruning},
  author={Wang, Junxiao and Guo, Song and Xie, Xin and Qi, Heng},
  booktitle={Proceedings of the ACM web conference 2022},
  pages={622--632},
  year={2022}
}

@article{romandini2024federated,
  title={Federated unlearning: A survey on methods, design guidelines, and evaluation metrics},
  author={Romandini, Nicol{\`o} and Mora, Alessio and Mazzocca, Carlo and Montanari, Rebecca and Bellavista, Paolo},
  journal={IEEE Transactions on Neural Networks and Learning Systems},
  year={2024},
  publisher={IEEE}
}

@misc{oasis_kaggle,
  author       = {Daithal, Nina},
  title        = {OASIS Alzheimer's Detection},
  howpublished = {Kaggle},
  year         = {2023},
  note         = {\url{https://www.kaggle.com/datasets/ninadaithal/imagesoasis}}
}

@article{hsu2019measuring,
  title={Measuring the effects of non-identical data distribution for federated visual classification},
  author={Hsu, Tzu-Ming Harry and Qi, Hang and Brown, Matthew},
  journal={arXiv preprint arXiv:1909.06335},
  year={2019}
}

@article{yang2023medmnist,
Author = {Yang, Jiancheng and Shi, Rui and Wei, Donglai and Liu, Zequan and Zhao,
   Lin and Ke, Bilian and Pfister, Hanspeter and Ni, Bingbing},
Title = {MedMNIST v2-A large-scale lightweight benchmark for 2D and 3D biomedical
   image classification},
Journal = {SCIENTIFIC DATA},
Year = {2023},
}

@article{gu2024unlearning,
  title={Unlearning during learning: An efficient federated machine unlearning method},
  author={Gu, Hanlin and Zhu, Gongxi and Zhang, Jie and Zhao, Xinyuan and Han, Yuxing and Fan, Lixin and Yang, Qiang},
  journal={arXiv preprint arXiv:2405.15474},
  year={2024}
}

@inproceedings{su2023asynchronous,
  title={Asynchronous federated unlearning},
  author={Su, Ningxin and Li, Baochun},
  booktitle={IEEE INFOCOM 2023-IEEE conference on computer communications},
  pages={1--10},
  year={2023},
  organization={IEEE}
}

@article{li2025achieving,
  title={Achieving Fair Medical Image Segmentation in Foundation Models with Adversarial Visual Prompt Tuning},
  author={Li, Yuqi and Li, Yanli and Zhang, Kai and Zhang, Fuyuan and Yang, Chuanguang and Guo, Zhongliang and Ding, Weiping and Huang, Tingwen},
  journal={Information Sciences},
  pages={122501},
  year={2025},
  publisher={Elsevier}
}

@inproceedings{li2025frequency,
  title={Frequency-aligned knowledge distillation for lightweight spatiotemporal forecasting},
  author={Li, Yuqi and Yang, Chuanguang and Zeng, Hansheng and Dong, Zeyu and An, Zhulin and Xu, Yongjun and Tian, Yingli and Wu, Hao},
  booktitle={Proceedings of the IEEE/CVF International Conference on Computer Vision},
  pages={7262--7272},
  year={2025}
}

@article{Li2025Efficient,
  author  = {Li, Yuqi and Zeng, Hansheng and Zhang, Fuyan and Yang, Chuanguang and Li, Yanli and Ding, Weiping},
  title   = {{Efficient Medical Image Segmentation via Reinforcement Learning-Driven K-Space Sampling}},
  journal = {IEEE Transactions on Emerging Topics in Computational Intelligence},
  year    = {2025},
  doi     = {10.1109/TETCI.2025.3621221},
  issn    = {2471-285X},
  publisher = {IEEE}
}

@article{li2025mmt,
  title={MMT-ARD: Multimodal Multi-Teacher Adversarial Distillation for Robust Vision-Language Models},
  author={Li, Yuqi and Dong, Junhao and Yang, Chuanguang and Wen, Shiping and Koniusz, Piotr and Huang, Tingwen and Tian, Yingli and Ong, Yew-Soon},
  journal={arXiv preprint arXiv:2511.17448},
  year={2025}
}

@article{li2025srkd,
  title={SRKD: Towards Efficient 3D Point Cloud Segmentation via Structure-and Relation-aware Knowledge Distillation},
  author={Li, Yuqi and Dong, Junhao and Dong, Zeyu and Yang, Chuanguang and An, Zhulin and Xu, Yongjun},
  journal={arXiv preprint arXiv:2506.17290},
  year={2025}
}

\end{document}